\documentclass[letterpaper, 10pt, conference]{ieeeconf}       

\IEEEoverridecommandlockouts                              
\usepackage{graphics} 
\graphicspath{{figures/}}
\usepackage{float}

\usepackage{epsfig} 
\usepackage{mathptmx} 
\usepackage{times} 
\usepackage{amsmath} 
\usepackage{amssymb} 
\usepackage{balance}

\usepackage[
  backgroundcolor=green,
  linecolor=green
  ]{todonotes}
\usepackage[nolist]{acronym}
\usepackage[ruled, linesnumbered, vlined]{algorithm2e}
\let\labelindent\relax
\usepackage[inline]{enumitem}

\newcommand{\del}{\partial}
\newcommand{\XX}{\mathcal{X}}

\usepackage{hyperref}
\hypersetup{
  colorlinks=true,
  linkcolor=black,
  citecolor=black,
  filecolor=black,
  urlcolor=black
}

\usepackage{tikz}
\usetikzlibrary{shadows}
\usepackage{xcolor}

\title{\LARGE \bf
An Optimal  Algorithm to Solve the \acl{CTAPF} Problem}

\author{Christian Henkel$^{1}$, Jannik Abbenseth$^{2}$ and Marc Toussaint$^{1}$
\thanks{$^{1}$Machine Learning and Robotics Lab, University of Stuttgart, Germany
        {\tt\small
       	post@henkelchristian.de,  marc.toussaint@\newline informatik.uni-stuttgart.de}}
\thanks{$^{2}$Fraunhofer Institute for Manufacturing
	Engineering and Automation IPA, Stuttgart, Germany
	{\tt\small jannik.abbenseth@\newline ipa.fraunhofer.de}}}%

\begin{document}

\begin{acronym}
    \acro{AGV}{Autonomous Guided Vehicle}
    \acro{CBS}{Conflict-Based Search}
    \acro{CDF}{Cumulative Distribution Function}
    \acro{CPPS}{cyber-physical production System}
    \acro{CPS}{cyber-physical system}
    \acro{CTAPF}{Combined Task Allocation and Path Finding}
    \acro{DCOP}{Distributed Constraint Optimization Problem}
    \acro{HEM}{Holistic Environment Model}
    \acro{HMM}{Hidden-Markov-Model}
    \acro{I4.0}{Industry 4.0}
    \acro{ICTS}{Increasing Cost Tree Search}
    \acro{ISV}{independent service vendor}
    \acro{LAN}{local area network}
    \acro{MAPF}{Multi-Agent Path Finding}
    \acro{MES}{Manufacturing Execution System}
    \acro{MATA}{Multi-Agent Task Allocation}
    \acro{MAPD}{Multi-Agent Pick-up and Delivery}
    \acro{MSB}{Manufacturing Service Bus}
    \acro{MILP}{Mixed Integer Linear Programming}
    \acro{MINLP}{Mixed Integer \textit{Non-Linear} Programming}
    \acro{PAAS}{platform as a service}
    \acro{PDF}{Probability Density Function}
    \acro{PDP}{Pickup and Delivery Problem}
    \acro{ROS}{Robot Operating System}
    \acro{SLAM}{Simultaneous Localization and Mapping}
    \acro{TAPF}{Combined Target-Assignment and Path-Finding}
    \acro{TCBS}{Task Conflict-Based Search}
    \acro{TPTS}{Token Passing with Task Swaps}
    \acro{UUID}{Universally Unique Identifier}
    \acro{VFK}{Virtual Fort Knox}
    \acro{WHCA*}{Windowed Hierarchical Cooperative A*}
    \acro{WLAN}{wireless \ac{LAN}}
\end{acronym}

\maketitle
\thispagestyle{empty}
\pagestyle{empty}

\begin{abstract}

  We consider multi-agent transport task problems where, e.g.\ in a factory setting, items have to be delivered from a given start to a goal pose while the delivering robots need to avoid collisions with each other on the floor.

  We introduce a \acf{TCBS} Algorithm to solve the combined delivery task allocation and multi-agent path planning problem optimally.
  The problem is known to be NP-hard and the optimal solver cannot scale. 
  However, we introduce it as a baseline to evaluate the sub-optimality of other approaches.
  We show experimental results that compare our solver with different sub-optimal ones in terms of regret.


\end{abstract}

\section{Introduction} 
In \acf{MAPF} the problem specification assigns a fixed start and goal state to each agent and the core challenge is to find collision free paths for all agents.
In contrast, both \acf{MATA} and \acf{MAPD} focus on solving the allocation of delivery jobs to the agents, while collisions between agent paths (e.g.\ in traffic networks) are not considered an issue.
This paper considers the joint problem of delivery task allocation and finding collision free paths for all agents jointly.
We denote this as the \acf{CTAPF} problem.

This problem setting is motivated by dense factory floor transportation systems.
Here items have to be transported from their current locations to goal locations by robotic agents that share narrow aisles and routes in the factory floor.
More generally, the problem setting captures any multi-agent domain where the tasks are to traverse from one state to another, and spatio-temporal constraints between the agents exist.

As our problem setting is a superset of multi-agent path finding, it is NP-hard. 
In this paper we introduce an explicit optimal solver for such problems, which necessarily cannot scale well with the number of agents.
We use the method to evaluate the regret (sub-optimality) of alternative approaches in our experiments.

In \autoref{sec:problem} we introduce the \ac{CTAPF} problem formulation.
We then present in \autoref{sec:solver} an optimal solver.
In \autoref{sec:evaluation} we show simulation experiments using our algorithm.

\section{Related Work}
The task allocation problem is a well researched challenge in the field of multi-agent systems and operations research.
We refer to it as \ac{MATA}.
A centralized approach to solve it is the Hungarian algorithm \cite{Kuhn1955}.
Further, there exist auction based approaches allowing to solve the problem decentralized \cite{Smith1980}.
More recently additional market-based approaches \cite{Dias2006}, reactive methods \cite{Parker1998} or biologically inspired ones \cite{Kube2000} have been introduced.
We model the problem in a way such that one agent can only perform one task at the time and every task only requires one agent.
In the taxonomy by \cite{Gerkey2004} this is an ST-SR-TA problem (Single-Task Robots, Single-Robot Tasks, Time-Extended Assignment) which is proven to be NP-hard \cite{Gerkey2004}.

The particular problem of transport task allocation is called \acf{PDP} \cite{Savelsbergh1995a} in operations research.
See \cite{Parragh2008} for a survey article.
It is defined by a set of agents and a number of requests of certain amount to be transported from one location to another.
The problem is often studied with time windows \cite{Lau2001} or to optimize vehicle capacity \cite{Taniguchi1999} especially in the \ac{AGV} domain \cite{Sinriech1991}.
Currently we do not consider time windows because they are usually not defined in industrial scenarios.
Also, we are concerned with the special case of the \ac{PDP} where every agent has a capacity of one unit, since this models the scenario best.
This may be also referred to as dial-a-ride problem \cite{Savelsbergh1995a}.

\acf{MAPF} is another intensively studied problem in multi-agent systems \cite{Allen1983}.
The decision in \ac{MAPF} concerns how a number of agents will be traveling to their goal poses without colliding, also an NP-hard problem \cite{Yu2013d}.
The colored pebble problem is comparable since the color of the pebble makes them not interchangeable \cite{Goraly}.
Solving the problem with collision avoidance at runtime can lead to deadlocks especially in narrow environments as discussed by \cite{Silver2005a} and more recently by \cite{Cap2016}.
Available sub-optimal solutions to the problem include Local Repair A* \cite{Zelinsky1992b}, WHCA* \cite{Silver2005a} and sampling based approaches like Multi-agent RRT* \cite{Cap2013a} and ARMO \cite{Kleiner2011a}.
Optimal solvers are \ac{ICTS} \cite{Sharon2013c} and \acf{CBS} \cite{Sharon2015}.
Our solution is based on the latter, because it can be extended with task-assignment and can then solve the introduced problem optimally.
Previously we elaborated \ac{CBS} with nonuniform costs in an industrial \ac{AGV} scenario \cite{Abbenseth2017}.

One problem formulation that is more closely related to transport systems is the \ac{TAPF} introduced by Ma and Koenig \cite{Ma2016a}.
It first solves the assignment problem and \ac{MAPF} problem but not concurrently, so the costs used for task allocation are not the true costs.
Instead of single goals we consider the whole transport task allocation.

A different type of problem is studied in vehicle routing with capacities \cite{Russell2005},
which focuses on the deliveries from one central depot based on certain demands.
This is a different problem in the sense that it considers only one origin and additionally considers capacity constraints.

The joint solution of \ac{MATA} and \ac{MAPF}, that we are proposing, was previously studied in \cite{Koes2005},
where the problem is considered as \ac{MILP} but due to collisions on the single-agent path level we think it should be considered a \ac{MINLP} Problem.
Therefore, the solver proposed by Koes et al. \cite{Koes2005} can not solve the problem optimally since it ignores agent-agent collisions.

A similar problem, by taking uncertainties into account, aims at applications in highly dynamic environments \cite{Woosley2013}.
The solution is also sub-optimal because agent-agent collisions are not considered at planning time.

Similar problems have recently also been formulated by Farinelli et al. \cite{Farinelli2016}, Ma et al. \cite{Ma2017} and Nguyen et al. \cite{Nguyen2017} for domains that do not involve transport tasks as we consider them.
Both find interesting sub-optimal solutions for \ac{TPTS} \cite{Ma2017}, based on \acp{DCOP} \cite{Farinelli2016} and using answer set programming (ASP) \cite{Nguyen2017}.
All of these approaches may solve the sub-problems optimally but they do not solve the combined problem optimally. 
This would require taking the implications that the task assignment makes into the path finding problem and vice versa.

The field of \textit{combined task and motion planning} \cite{Wolfe2010} is also combining two planning domains but for mobile manipulators.
Our algorithm borrows the concept of hierarchical planners and the reaction to planning errors i.e. when no path is found.

In the literature we can find many approaches that go towards solving the problems of \ac{MATA} and \ac{MAPF} in interesting settings.
Little research is done in combining both which is the core aspect addressed in this paper.
To our knowledge it is the first to solve the combined \ac{MATA} and \ac{MAPF} problem optimally.



\section{Problem Formulation} \label{sec:problem}
As discussed above, \ac{MAPF} and \ac{MATA} have been researched intensively. We now propose a combined problem formulation.

We consider a configuration space $\XX$, and $n$ agents with joint configuration state $x = (x_1,..,x_n) \in \XX^n$.

We have a set $t = \{t_1, t_2, ..., t_m\}$ of $m$ tasks, each one consists of a start $x^s_i$ and a goal $x^g_i$ configuration, $t_i = (x^s_i, x^g_i) \in \XX^2$.

Once an agent visits a task's start pose, it is automatically assigned to the task, and we say the task is \textit{running}.
It is fixed to this task and can not be reassigned to another task before reaching its goal.

The problem is to find a minimum duration $T \in \mathbf{N}$ and paths $P_j$ of length $T$ for each agent $j$ such that every task is fulfilled.
In the following we more formally define the constraints on the path (that they follow a roadmap and do not collide), and when tasks are fulfilled.

Each path $P_j = (P_j^0,..,P_j^T)$ is a sequence of $T+1$ configurations, where $P_j^0\in\XX$ is constrained to be the start configuration of agent $j$, and $P_j^{t+1} \in \del(P_j^t) \subseteq \XX$ is adjacent to the agents previous configuration $P_j^t$. Adjacency is defined by a roadmap $\del(\bullet)$ which, for each $x\in\XX$ defines its neighborhood $\del(x)\subseteq\XX$.

Given the set of paths $\{P_1,..,P_n\}$ we require them to be non-colliding, meaning that $\nexists t,j,k: P_j^t = P_k^t$.
In addition, we also require agent movements to be non-colliding, that is, they must not swap places on the roadmap, $\nexists t,j,k: P_j^{t+1} = P_k^t \wedge P_j^t = P_k^{t+1}$.

Given a path $P_j$ of an agent $j$, we can compute the implicit assignment $\tau_j^T \in \{1,..,m\}$ of agent $j$ to a task $i\in\{1,..,m\}$ at time step $T$.
The mapping from $P_j$ to $\tau_j$ is unique given the rule that a task becomes automatically active if the agent visits its start configuration and ends only when it reaches the task goal.
Based on this it is also uniquely defined whether a task $i$ is fulfilled by a path $P_j$. Therefore, given all agent paths $P_j$ we can evaluate whether all tasks are fulfilled.

In summary, the input to the problem are the agent's initial configurations $\{P_1^0,..,P_n^0\}$, the roadmap defined by $\del(\bullet)$, and the set of tasks $t$. 
The output is the total time $T$ per task and all agent paths $\{P_1,..,P_n\}$, subject to the path constraints and fulfillment of all tasks. 
The objective is to minimize $\sum_{i\in(0, .. m)} T_i$, the sum of duration per tasks until it is fulfilled.

In the evaluation, we use a grid-map as roadmap graph.
The approach can work on any roadmap graph that fulfills the property that none of the poses in the graphs vertices are in collision with any other.

Optimality is only guarantied with respect to the chosen roadmap.
I. e. the best solution duration is measured based on the number of roadmap edges traveled in that given roadmap.
Therefore, the duration in actual time units depends on the length of the edges and speeds of the agents.

\section{Optimal Solver} \label{sec:solver}
\subsection{Concept}
Our \acf{TCBS} searches on the level of task assignments for all agents. Given a configuration of task assignments and \emph{neglecting collision constraints}, we can use standard single-agent path finding to compute the corresponding optimal paths that connect the agents configurations to the start \emph{and} the start to the goal configurations of all assigned tasks.
To also account for the path collision constraints we follow the approach of conflict-based search.
This introduces additional decision variables in the search tree that represent the ``decision'' to put an explicit avoid constraint on an agent path.
That is, we not only search for task assignments but also avoid configurations that would contain a collision. 
We will discuss below that this approach does not compromise optimality of the method.

To give a first intuition on how the search tree is built, we describe it briefly before giving the technical description afterwards:
If there are no currently running tasks, we start with an empty assignment in the root node.
Therefore, at the beginning all tasks and all agents are unassigned.
At every node we perform \textit{one} of the two following types of expansion:
\begin{enumerate*}[label={\alph*)}]
	\item \emph{Iff no collision between agents exists}:
	Add children for all possible combinations of one still unassigned tasks to one agent.
	This means that the task will be added at the end of the agents task list.
	\item \emph{else (i.e. iff a collision between agents is detected)}:
	The expansion adds nodes where the configuration in the roadmap (or the motion between two configurations) is avoided at the time of the collision, one for each colliding agent.
\end{enumerate*}
See also \autoref{fig:tree}.


\subsection{Decision variables} \label{ssec:variables}
The discrete decision variables define node $s$ of the search tree, where $s.\tau$ defines the tasks assigned to agents and $s.\beta$ contains avoided configurations (or the motion between two configurations).

For each agent $j$, one assignment list $\tau_j, j \in \{1, ..., n\}$ exists.
Each assignment list $\tau_j = (t_{j 0}, t_{j 1}, ...)$ defines the sequence of tasks that the agent $j$ has to fulfill \textit{in order}.
Where each list element $t_{j k} \in \{1,...,m\}, k \in \mathbf{N}$ references a task.

To establish collision avoidance we define configurations or motions between configurations that an agent avoids at a given time.
For each agent $j$, one avoid list $\beta, j \in \{1, ..., n\}$ is defined.
Each list can contain multiple avoided configurations: $\beta_j = (\beta_{j 0}, \beta_{j 1}, ...)$.
And one avoided configuration is defined by \textit{either} $\beta_{j k} = (x_\beta, T_\beta), k \in \mathbf{N}$,
where $x_\beta$ is the configuration to be avoided at time $T_\beta$.
\textit{Or} $\beta_{j 0} = (x_{\beta 0}, x_{\beta 1}, T_\beta)$,
where the motion between $x_{\beta 0}$ and $x_{\beta 1}$ is avoided at time $T_\beta$.

\subsection{Search Algorithm} \label{ssec:a-star}
Given the specific problem formulation in \autoref{sec:problem} we can now describe an optimal solver as A* tree search over assignments of these decision variables.
We perform this search by using
\begin{itemize}
  \item Algorithm~\ref{alg:expand} to expand nodes,
  \item Algorithm~\ref{alg:goal} the goal test,
  \item Algorithm~\ref{alg:heuristic} to decide which nodes to expand first and
  \item Algorithm~\ref{alg:cost} to determine the cost-so-far function for a node
\end{itemize}

\begin{figure}
\vspace{5pt}  
\definecolor{bmhred}{RGB}{71,21,31}
\begin{tikzpicture}[level distance = 2.5em, sibling distance=4.3em, scale=0.85,
  every node/.style = {shape=rectangle,
    draw, align=center, font=\sffamily,
    text centered, anchor=north,
    top color=white, drop shadow,
    scale = 0.62}]]
  \node [sibling distance=3.5em, bottom color={rgb:red,35;green,62;blue,78}] {$s.\tau = (\{\}, \{\})$\\$s.\beta = (\{\}, \{\})$}
    child { [bottom color={rgb:red,35;green,62;blue,78}] node {$(\{\mathbf{0}\}, \{\})$\\$(\{\}, \{\})$} }
    child { [sibling distance=4.5em, bottom color={rgb:red,35;green,62;blue,78}] node {$(\{\mathbf{1}\}, \{\})$\\$(\{\}, \{\})$}
      child { [] node {$(\{1, \mathbf{0}\}, \{\})$\\$(\{\}, \{\})$} }
      child { [] node {$(\{1, \mathbf{2}\}, \{\})$\\$(\{\}, \{\})$} }
      child { [] node {$(\{1\}, \{\mathbf{0}\})$\\$(\{\}, \{\})$} }
      child { [sibling distance=5.5em] node {$(\{1\}, \{\mathbf{2}\})$\\$(\color{bmhred}\mathbf{collision}\color{black})$}
        child { [bottom color={rgb:red,71;green,21;blue,31}] node {$(\{1\}, \{2\})$\\$(\{\mathbf{(5, 3, 6)}\}, \{\})$}
          child { [bottom color={rgb:red,41;green,56;blue,29}, very thick] node {$(\{1, \mathbf{0}\}, \{2\})$\\$(\{(5, 3, 6)\}, \{\})$} }
          child { [bottom color={rgb:red,35;green,62;blue,78}] node {$(\{1\}, \{2, \mathbf{0}\})$\\$(\{(5, 3, 6)\}, \{\})$} }
        }
        child { [bottom color={rgb:red,71;green,21;blue,31}] node {$(\{1\}, \{2\})$\\$(\{\}, \{\mathbf{(5, 3, 6)}\})$} }
      }
    }
    child { [bottom color={rgb:red,35;green,62;blue,78}] node {$(\{\mathbf{2}\}, \{\})$\\$(\{\}, \{\})$} }
    child { [bottom color={rgb:red,35;green,62;blue,78}] node {$(\{\}, \{\mathbf{0}\})$\\$(\{\}, \{\})$} }
    child { [bottom color={rgb:red,35;green,62;blue,78}] node {$(\{\}, \{\mathbf{1}\})$\\$(\{\}, \{\})$} }
    child { [bottom color={rgb:red,35;green,62;blue,78}] node {$(\{\}, \{\mathbf{2}\})$\\$(\{\}, \{\})$} };
\end{tikzpicture}
\caption{Example of a \ac{TCBS} tree for $n=2$, $m=3$. The second level shows the big branching factor which is in this case 6. The leftmost node has for example a configuration, where the first agent executes task 0. In the third displayed level, the assignments are extended further so that the leftmost node indicates a configuration where the first agent would execute task 1 and then 0.
In the rightmost node occurs a collision which is solved through avoiding the configuration (here cell 	5, 3 at time 6) in the fourth level (red nodes). The last unassigned task 0 is assigned to the first agent which leads to a solution in the left leaf node of the final level.} \label{fig:tree} 
\end{figure}
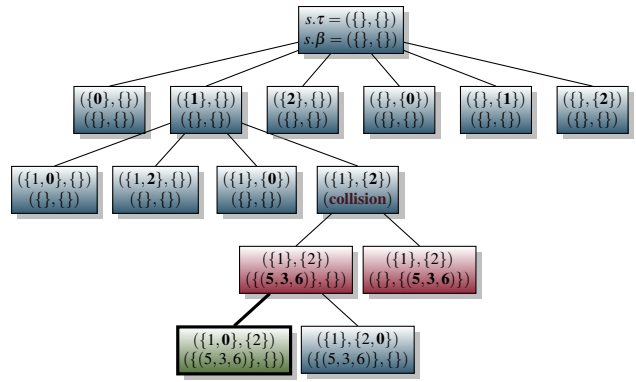

\subsection{Expand} \label{ssec:expand}
\begin{algorithm}[!t]
  \small
  \KwIn{node $s$}
  \textsf{children} = []\\
  \If{$\textsf{collision}$ in $s$}{
  	// for all agents in collision\\
    \For{$j~in~\textsf{collision}$}{
      // add avoided configuration (or motion between them) at a certain point in time for agent $j$\\
      \textsf{children} $\gets$ \textsf{children} + $(s.\beta_j + (x_\beta, T_\beta))$
    }
  }
  \Else{
  	// for all unassigned tasks\\
    \For{$t \notin s.\tau$}{
      // for all agents\\
      \For{$j$}{
        \textsf{children} $\gets$ \textsf{children} + $(s.\tau_j + t)$
      }
    }
  }
  \Return{\textsf{children}}\\

  \caption{Expand}
  \label{alg:expand}
\end{algorithm}

In the tree search we use \autoref{alg:expand} to expand a node.
\autoref{fig:tree} shows how a search tree after multiple expansions may look like.
It is especially visible how the two different types of expansion shape the tree.
The blue nodes are added by an assignment expansion while the red ones are added to resolve a collision.

\subsection{Goal Test} \label{ssec:goal}
\begin{algorithm}[t]
	\small
	\KwIn{node $s$}
	// if in collision\\
	\If{$\textsf{collision}$ in $s$}{
		\Return{\textsf{false}}
	}
	// or if unassigned tasks exist\\
	\ElseIf{$\exists t \notin s.\tau$}{
		\Return{\textsf{false}}
	}
	// otherwise it is the goal\\
	\Else{
		\Return{\textsf{true}}\\
	}

	\caption{Goal Test}
	\label{alg:goal}
\end{algorithm}

\autoref{alg:goal} shows the algorithm to do the goal test that is performed on any evaluated node.
It requires the node to have no collisions and all tasks to be assigned.

\subsection{Heuristic} \label{ssec:heuristic}
\SetKwFunction{KwDist}{Distance}
\begin{algorithm}[t]
  \small
  \KwIn{node $s$}
  \textsf{cost} = 0\\
  // for all unassigned tasks\\
  \For{$t \notin s.\tau$}{
	  \textsf{cost} $\gets$ \textsf{cost} + \KwDist(closest\_agent, start($t$)) \\
	  \textsf{cost} $\gets$ \textsf{cost} + \KwDist(start($t$), goal($t$))
  }
  \Return{\textsf{cost}}
  \caption{Heuristic}
  \label{alg:heuristic}
\end{algorithm}

The heuristic function ($h()$ function of A*) is presented in \autoref{alg:heuristic}.
It evaluates the cost of the best-possible assignment. 
This sums for each unsigned task the distance to the closest agent which might be either its starting position or the goal pose of a previously executed task.

\subsection{Cost-so-far} \label{ssec:cost}
\SetKwFunction{KwPaths}{getPaths}
\begin{algorithm}[t]
  \small
  \KwIn{node $s$}
  \textsf{cost} = 0\\
  // Plan single-agent paths according to assignments and avoided configurations:\\
  P $\gets$ \KwPaths(s)
  \For{$p_x \in P$}{
    \If{$p_j \equiv p_x$}{
      // Time at task end added to cost:\\
      \textsf{cost} $\gets$ \textsf{cost} + $p_j.T_{final}$ \nllabel{alg:cost:T} \\
    }
  }
  \If{collisions in $P$}{
    $s.\textit{\textsf{collision}}$ $\gets$ \textsf{true}
  }d
  \Return{$s$, \textsf{cost}}
  \caption{Cost-so-far}
  \label{alg:cost}
\end{algorithm}

The cost function ($g()$ function of A*) in \autoref{alg:cost} evaluates the cost of the node.
A sum of all arrival times is used as cost.
This method is also used to evaluate if collisions are present in a node by computing the single-agent paths.
If a collision is detected, we still calculate costs. 
It allows for the freedom to solve collision by changing the task assignment which is a benefit of the \ac{CTAPF} problem formulation.

\subsection{Path Function} \label{ssec:path_func}
Our approach requires a function to compute a set of paths for all agents for any given node.
This plans single agent paths according to the assigned tasks and the configurations (or motions between two configurations) to be avoided.
We also use A* here, but on the environments roadmap.
Additionally we \textit{cache} previously calculated paths to use it multiple times.
Both of the presented algorithms (Alg.~\ref{alg:heuristic} and Alg.~\ref{alg:cost}) use the notion of $\mathtt{Distance(a,b)}$ which is a method that returns the Manhattan Distance between two points or, if present in the previously mentioned cache, returns the actual distance of the shortest path.

\subsection{Optimality} \label{ssec:optimality}
An A* search algorithm is considered optimal if the heuristic function is admissible \cite{Pearl1984}:
\begin{equation}
  h(s) \leq h^*(s) ~ \forall ~ s
\end{equation}
Where $h(s)$ is the heuristic of node $s$ while $h^*(s)$ is the optimal cost.

\begin{figure*}[!ht]
	\centering
	\includegraphics[width=.9\textwidth]{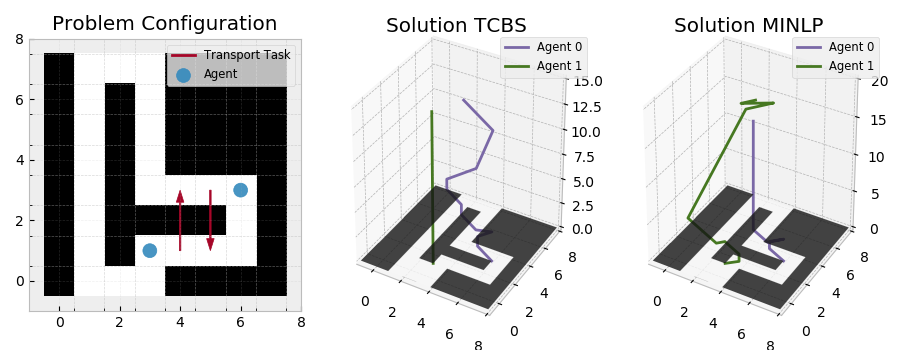}
	\caption[]{Example problem configuration (left) and solution as path set in space-time (with time as vertical dimension) by our \ac{TCBS} planner (middle) and by separately solving task allocation as \ac{MINLP} optimization problem and path finding with \ac{CBS} (right). For an animated version see video\footnotemark.}
	\label{fig:scenario}
\end{figure*}
\footnotetext{\url{https://ct2034.github.io/miriam/iros2019/video/}}
The heuristic (\autoref{alg:heuristic}) is optimistic since it considers the distance of the closest agent to the tasks goal.
This is never greater than the actual path distance:
\begin{equation}
  \begin{aligned}
    \mathtt{Distance}(closest\_agent, start(task)) \\
    \leq
    \mathtt{Path}(assigned\_agent, start(task)).T_{final}
  \end{aligned}
\end{equation}
Where $T_{final}$ indicates the time at the end of the path assuming constant motion speed.

And for the length of a transport task, the optimality criterion is the same as for single-agent path finding:
\begin{equation}
  \begin{aligned}
    \mathtt{Distance}(start(task), goal(task)) \\
    \leq
    \mathtt{Path}(start(task), goal(task)).T_{final}
  \end{aligned}
\end{equation}
The methods $\mathtt{Distance(a,b)}$ and $\mathtt{Path(a,b)}$ are described in \autoref{ssec:path_func} in more detail.

The consistency of the heuristic would be required, if we were searching on a general graph with no tree structure \cite{Russell2010}.
Since \autoref{alg:expand} only expands nodes by adding information, the search graph \textit{has} tree structure and consistency is no required criterion.

That is, our approach adds avoided configurations does not compromise the optimality.
The proof is available in the original publication of \ac{CBS}\cite{Sharon2015}.

\section{Evaluation} \label{sec:evaluation}
The source code of our implementation in Python is available online\footnote{\url{https://ct2034.github.io/miriam/iros2019}}.

\begin{figure}[]
	\centering
	\includegraphics[width=.747\columnwidth]{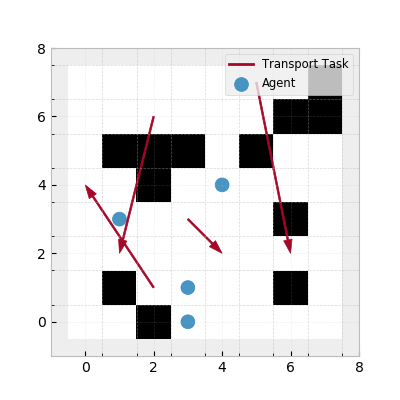}
	\caption{One random configuration that was used for the comparison of different algorithms. Indicating agent locations and transport tasks. Black squares in the background show obstacles.}
	\label{fig:random}
\end{figure}
\begin{figure}[]
	\centering
	\includegraphics[width=.8\columnwidth]{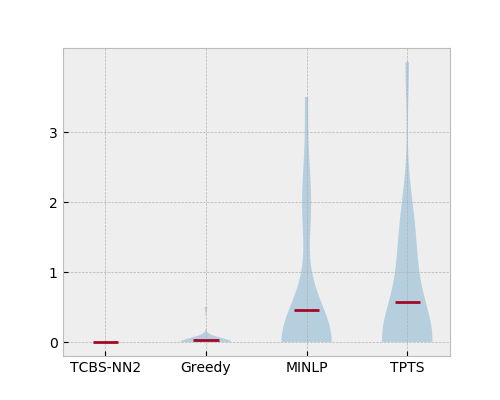}
	\caption{Task duration surplus (i.e. regret) for the indicated algorithms in comparison to the optimal solution of the same problem computed by \ac{TCBS}. Violin plot with red lines indicating the mean value.}
	\label{fig:solutionquality}
\end{figure}
\begin{figure*}[]
	\centering
	\includegraphics[width=\textwidth]{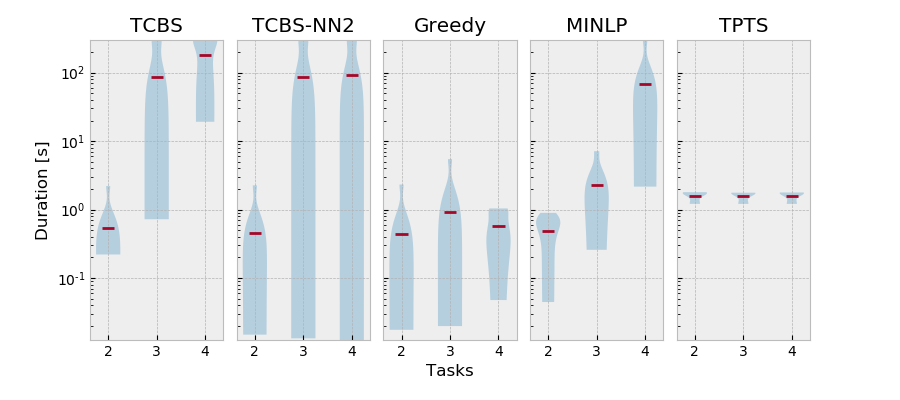}
	\caption{Planning time required by the different algorithms for different problem sizes (i.e. number of tasks). Violin plot with red lines indicating the mean value.}
	\label{fig:duration}
\end{figure*}

\subsection{Example Problem and Solution}
\autoref{fig:scenario} shows an example of one problem configuration and the resulting path set after solving the problem.
We can compare the two solutions by \ac{TCBS} (middle) and \ac{MINLP} \cite{Koes2005} (right).
The algorithms received a set of agent positions and tasks (\autoref{sec:problem}) as input and returned a set of paths for the agents.
From the paths in the pictures follow the task execution times displayed in \autoref{tab:example} per task.
We compare the total time for all tasks $T$, as discussed in \autoref{sec:problem}.

\begin{table}[h]
\center
\begin{tabular}{l*{6}{c}r}
Algorithm         & Task 1 & Task 2 & Total \\
\hline
\ac{TCBS}         & 6 & 15 & 21  \\  
\ac{MINLP}        & 6 & 20 & 26  \\
\end{tabular}
\caption{Execution times for the tasks in the example.}
\label{tab:example}
\end{table}

It is visible that \ac{TCBS} produces a solution quality superior to \ac{MINLP} while it does only utilize one agent.
The task allocation step of \ac{MINLP} assigns each agent to the closest task.
When the tasks are executed, the two agents in \ac{MINLP} block each other.
In this sense the \ac{TCBS} algorithm successfully finds a solution that takes the transport task execution into account.
This shows how special cases require to jointly solve the \ac{MATA} and the \ac{MAPF} problems which is the main claim of this paper.

\subsection{Algorithms for Comparison} \label{ssec:algorithmscompare}
To give further insight into the different sub-optimal solvers we are introducing them in the following before evaluating them afterwards.

\begin{itemize}
\item \textbf{\ac{TCBS}} The optimal planner described in this paper.

\item \textbf{\ac{TCBS}-NN2} (\ac{TCBS} for 2 nearest neighbors) A version of our algorithm which uses a sub-optimal nearest-neighbor search \cite{Muja2014} to only consider the n (here $n = 2$) closest of all possible assignments.
Additionally, it uses sub-optimal heuristics to resolve collision faster. It is still not scalable but can find solutions for much bigger problems than \ac{TCBS}

\item \textbf{Greedy} A simple local search algorithm that incrementally assigns tasks to agents using a sub-optimal nearest neighbor search.
It finds the currently closest pair of unassigned agents and tasks \textit{or} tasks and tasks for consecutive execution.
See \autoref{alg:greedy} for details.
Then we solve \ac{MAPF} by the \ac{CBS} algorithm.

\item \textbf{\ac{MINLP}} The separate solution of assignment and path finding problem proposed by \cite{Koes2005} has been implemented using the Bonmin solver \cite{Bonami2005} for the assignment problem and \ac{CBS} for path finding.

\item \textbf{\ac{TPTS}} The original goal of this approach is the lifelong version of \ac{MAPF}.
The task assignment is solved in advanced but optimized by swapping of tasks \cite{Ma2017}.
The implementation was provided by the authors of the paper and is available online\footnote{\url{http://github.com/ct2034/cobra}}.

\begin{algorithm}[ht]
  \small
  \KwIn{agents, tasks}
  free\_agents = agents.copy()

  free\_tasks = tasks.copy()

  consecutive = dict()

  agent\_task = dict()

  \While{len(free\_tasks) $>$ 0}{
    \If{len(free\_tasks) $>$ len(free\_agents)}{
      poses = free\_agents $\cup$ task\_ends
    }
    \Else{
      poses = free\_agents
    }
    closest\_pose, closest\_task = nearest\_neighbor(poses, free\_tasks)

    free\_tasks.remove(closest\_task)

    \If{type(closest\_pose) == task}{
      consec[task] = closest\_pose // (task)
    }
    \Else{
      agent\_task[task] = closest\_pose

      free\_agents.remove(closest\_pose) // as agent
    }
  }
  \Return{consecutive, agent\_task}

  \caption{Greedy Assignment}
  \label{alg:greedy}
\end{algorithm}
\end{itemize}
\pagebreak
\subsection{Solution Regret}
While the previous example was designed to explicitly show the potential shortcomings of the separate solution of \ac{MATA} and \ac{MAPF} problem, we want to focus on a wider comparison of regret for the available approaches now.
To allow a more general evaluation, we use now randomized scenarios for this comparison.

This uses a 8x8 map filled with 20\% random obstacles each scenario receives 2, 3 or 4 random tasks.
Per configuration 30 trials of different random configurations were tested.
An example of one randomly generate scenario is shown in \autoref{fig:random}.

\autoref{fig:solutionquality} shows the mean value of regret for different solvers.
This is, how much more duration per task was required in the respective solution \textit{in comparison to the optimal solution}.
It can be seen from the figure that most planners have a low mean value, meaning they found the optimal solution most of the time.
But different planners show different distributions of outliers in the data.
No deviation is visible for TCBS-NN2 which is the only algorithm in this comparison that solves \ac{MATA} and \ac{MAPF} in combination.
It shows that the joint problem solving delivers good overall solution quality (i.e. small regret).

The greedy algorithm shows also a very small regret, demonstrating that also simple approaches can lead to small regret.

Both \ac{MINLP} and \ac{TPTS} show the ability to find the optimal solution in the mean case as well but especially \ac{TPTS} has a significant amount of data with higher regret.
This demonstrate that \emph{traditional} separate solving of \ac{MATA} and \ac{MAPF} can lead to a considerable regret.

\subsection{Planning Time}
To compare the algorithms towards their practical usability, we evaluate their planning duration.
We used the same random scenarios as introduced before and for this experiment we also differentiate between problem size to elaborate scaling effects in the duration.
Problem size is defined as the number of tasks that are introduced.

\autoref{fig:duration} shows the planning time for the approaches.
Compared are all the planners mentioned in \autoref{ssec:algorithmscompare}
It is clearly visible that our \ac{TCBS} algorithm does not scale well with the problem size.
This is due to the problem being NP-hard.
\pagebreak
The results for TCBS-NN2 show similar scaling effect but also indicate a tendency towards faster computations times for larger problem sizes.
This is due to the algorithms pruning of the search tree by nearest neighbor search.

The greedy algorithm shows no visible scaling and fastest mean performance in this comparison.
It is very efficient because it always assigns the closest agent.

The \ac{MINLP} algorithm has a better overall performance than \ac{TCBS} and  TCBS-NN2 but shows exponential scaling as well.
This is due to the optimization problems complexity also being NP-hard.

Results for \ac{TPTS} show good performance and also scales well.
The method of task swaps lead to this results because it is independent of the problem size.

Both \ac{TPTS} and the greedy algorithm could therefore be taken into consideration for practical use although they lack solution quality as seen in the previous experiment.

\section{Conclusion} \label{sec:concludions}
\acf{CTAPF} is a practically highly relevant problem and has, to our knowledge, not been solved optimally previously.
We provided a detailed formalization of the problem.
The problem can be solved optimally by the proposed \acf{TCBS} algorithm based on a dynamically built search-tree.
This search-tree incrementally assigns tasks to agents and resolves collisions between agents as they occur.
In comparison to other approaches is demonstrated how the combined solving of task allocation and path finding improves the solution quality over solutions that solve them separately.
Our optimal algorithm is intended to support the development of sub-optimal planners by evaluating their regret. 
In addition, we proposed an approximate version of the solver with improved scaling which demonstrates smaller regret than the competing methods.


\balance
\bibliographystyle{IEEEtran}
\bibliography{lit}

\end{document}